\documentclass[10pt, a4paper]{article}
\usepackage{lrec}
\usepackage{graphicx}
\usepackage{tabularx}
\usepackage{balance}
\usepackage{soul}
\usepackage{epstopdf}
\usepackage[latin1]{inputenc}

\usepackage{hyperref}
\usepackage{xstring}
\title{EDA: Enriching Emotional Dialogue Acts using an Ensemble of \\Neural Annotators}
\name{Chandrakant Bothe, Cornelius Weber, Sven Magg, and Stefan Wermter}

\address{
Knowledge Technology, Department of Informatics,
University of Hamburg, \\
Vogt-Koelln-Str. 30, 22527 Hamburg, Germany \\
\url{www.informatik.uni-hamburg.de/WTM/} \\
\{bothe,weber,magg,wermter\}@informatik.uni-hamburg.de
}

\abstract{
The recognition of emotion and dialogue acts enriches conversational analysis and help to build natural dialogue systems.
Emotion interpretation makes us understand feelings and dialogue acts reflect the intentions and performative functions in the utterances.
However, most of the textual and multi-modal conversational emotion corpora contain only emotion labels but not dialogue acts.
To address this problem, we propose to use a pool of various recurrent neural models trained on a dialogue act corpus, with and without context. 
These neural models annotate the emotion corpora with dialogue act labels, and an ensemble annotator extracts the final dialogue act label.
We annotated two accessible multi-modal emotion corpora: IEMOCAP and MELD. 
We analyzed the co-occurrence of emotion and dialogue act labels and discovered specific relations.
For example, \textit{Accept/Agree} dialogue acts often occur with the \textit{Joy} emotion, 
\textit{Apology} with \textit{Sadness}, and
\textit{Thanking} with \textit{Joy}.
We make the Emotional Dialogue Acts (EDA) corpus publicly available to the research community for further study and analysis.
\\ \newline }

\begin{document}

\maketitleabstract

\section{Introduction}

With the growing demand for human-computer/robot interaction systems, detecting the emotional state of the user can substantially benefit a conversational agent to respond at an appropriate emotional level. 
Emotion recognition in conversations has proven valuable for potential applications such as response recommendation or generation, emotion-based text-to-speech, personalization. 
Human emotional states can be expressed verbally and non-verbally \cite{ekman1987universalemos,osgood1975cross}.
However, while building an interactive dialogue system, the interface needs dialogue acts. 
A typical dialogue system consists of a language understanding module which requires to determine the meaning and intention in the human input utterances \cite{wermter1996learning,berg2015nadia,ultes2017pydial}.
Also, in discourse or conversational analysis, dialogue acts are the main linguistic features to consider \cite{bothe2018interspeech}. 
A dialogue act provides an intention and performative function in an utterance of the dialogue.
For example, it can infer a user's intention by distinguishing \textit{Question}, \textit{Answer}, \textit{Request}, \textit{Agree/Reject}, etc. and performative functions such as \textit{Acknowledgement}, \textit{Conversational-opening or -closing}, \textit{Thanking}, etc.
% Emotion state is contextualize with dialogue act; GradAccent
The dialogue act information together with emotional states can be very useful for a spoken dialogue system to produce natural interaction \cite{ihasz2018emotions}.

The research in emotion recognition is growing, and many datasets are available, such as text-, speech- or vision-based, and multi-modal-based emotion data.
Emotion expression recognition is a challenging task, and hence multimodality is crucial \cite{ekman1987universalemos}.
However, few conversational multi-modal emotion recognition datasets are available, for example, IEMOCAP \cite{busso2008iemocap} or SEMAINE \cite{McKeown2012SEMAINE}, MELD \cite{poria2019meld}.
They are multi-modal dyadic conversational datasets containing audio-visual and conversational transcripts.
Every utterance in these datasets is labelled with an emotion label.

\begin{figure*}[!t]
\begin{center}
%\fbox{\parbox{6cm}{
%This is a figure with a caption.}}
\includegraphics[width=0.95\linewidth]{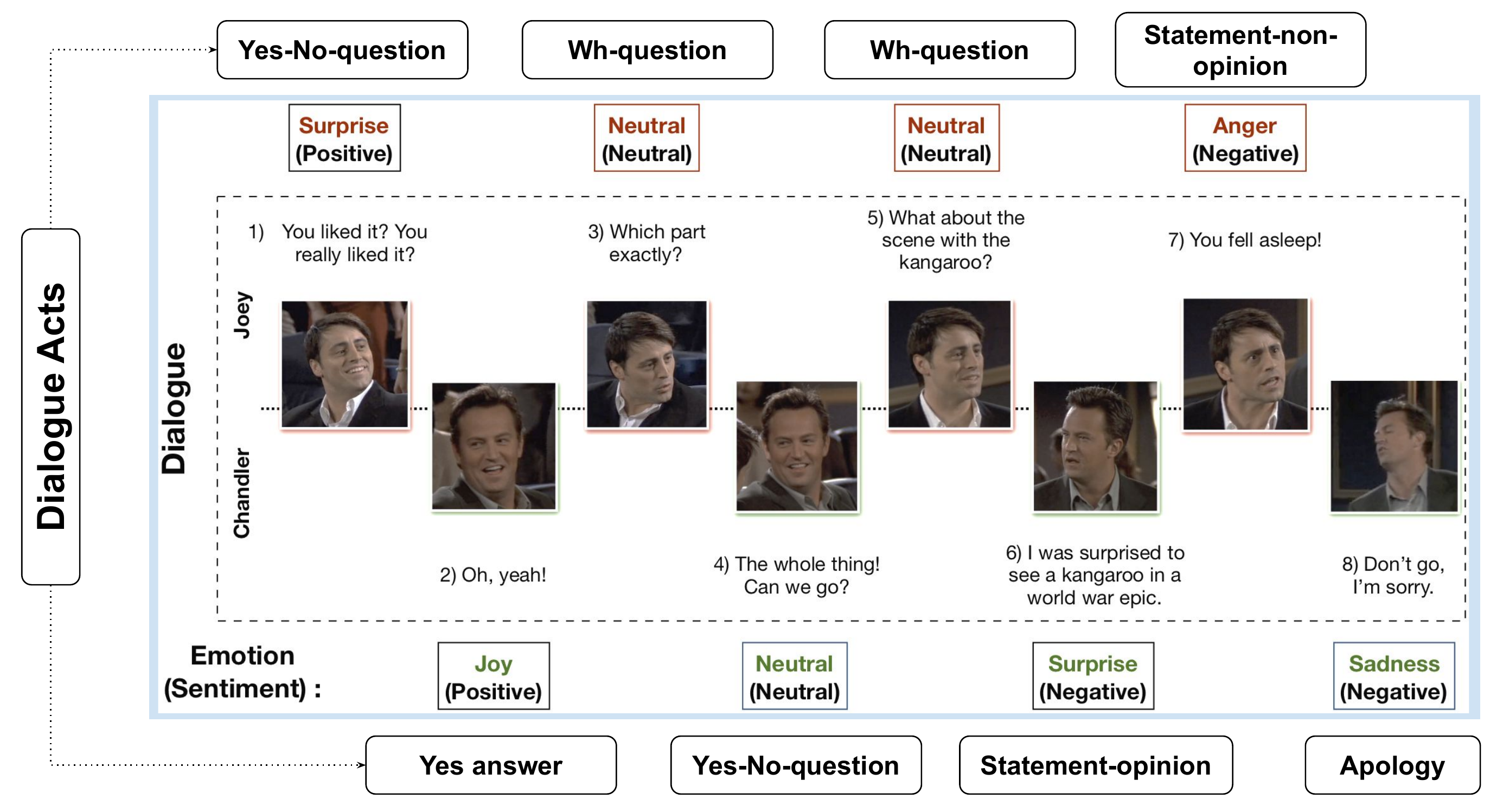}
\caption{Emotional Dialogue Acts: Example of a dialogue from MELD representing emotions and sentiment (rectangular boxes), in our work, we add dialogue acts (rounded boxes). Image source Poria et al. (2019). 
%TODO: central image is from MELD original paper; if necessary to change? 
}
\label{fig:dialogue-da-emotion}
\end{center}
\end{figure*}

In our research here, we propose an automated neural ensemble annotation process for dialogue act labelling.
% and final labels are observed to correct by a human annotator.
Several neural models are trained with the Switchboard Dialogue Act (SwDA) corpus \cite{godfrey1992switchboard,shribergSwitch} and used for inferring dialogue acts on the emotion corpora.
We integrate five model output labels by checking majority occurrences (most of the model labels are the same) and ranking confidence values of the models. 
% In the end, human annotator, from the authors ourselves, looks at the final labels and corrects if missed by the pool of automated annotators.
%The annotation process is detailed in Section \ref{section:annotation-process}
We have annotated two potential multi-modal conversation datasets for emotion recognition: IEMOCAP (Interactive Emotional dyadic MOtion CAPture database) \cite{busso2008iemocap} and MELD (Multimodal EmotionLines Dataset) \cite{poria2019meld}.
Figure \ref{fig:dialogue-da-emotion}, shows an example of the dialogue act tags with emotion and sentiment labels from the MELD corpus and we confirmed the reliability of annotations with inter-annotator metrics.
We analyzed the co-occurrences of the dialogue act and emotion labels and discovered an essential relationship between them: individual dialogue acts of the utterances show significant and useful association with corresponding emotional states.
For example, the \textit{Accept/Agree} dialogue act often occurs with the \textit{Joy} emotion while \textit{Reject} with \textit{Anger}, \textit{Acknowledgements} with \textit{Surprise}, \textit{Thanking} with \textit{Joy}, and \textit{Apology} with \textit{Sadness}, etc.
The detailed analysis of the emotional dialogue acts (EDAs) and annotated datasets are being made available at the Knowledge Technology website\footnote{\url{www.inf.uni-hamburg.de/en/inst/ab/wtm/research/corpora}
IEMOCAP (\url{https://sail.usc.edu/iemocap}) is available only with speaker IDs.}.

% TODO: small diagram to introduce the concept

\section{Annotation of Emotional Dialogue Acts}
\label{section:annotation-process}

\subsection{Data for Conversational Emotion Analysis}

There are two emotion taxonomies: (1) discrete emotion categories (DEC) and (2) fined-grained dimensional basis of emotion states (DBE).
The DECs are Joy, Sadness, Fear, Surprise, Disgust, Anger and Neutral as identified by Ekman et al. \shortcite{ekman1987universalemos}.
The DBE of the emotion is usually elicited from two or three dimensions \cite{osgood1975cross,russell1977evidence,cowie2003describing}.
A two-dimensional model is commonly used with Valence and Arousal (also called activation), and in the three-dimensional model, the third dimension is Dominance.
The IEMOCAP dataset is annotated with all DECs and two additional emotion classes, Frustration and Excited.
% It also has neutral emotion class but accompanied by non-neutral (xxx) class.
The IEMOCAP dataset is also annotated with three DBE, that includes Valance, Arousal and Dominance \cite{busso2008iemocap}.
The MELD dataset \cite{poria2019meld}, which is an evolved version of the Emotionlines dataset developed by \cite{chun2018emotion_lines}, is annotated with exactly 7 DECs and sentiments (positive, negative and neutral).

\subsection{Dialogue Act Tagset and SwDA Corpus}

There have been different taxonomies for dialogue acts: speech acts \cite{austin1962things} refer to the utterance, not only to present information but to the action is performed.
Speech acts were later modified into five classes (Assertive, Directive, Commissive, Expressive, Declarative) \cite{searle1979}.
There are many such standard taxonomies and schemes to annotate conversational data, and most of them follow the discourse compositionality. 
These schemes have proven their importance for discourse or conversational analysis \cite{skantze2007}.
During the increased development of dialogue systems and discourse analysis, the standard taxonomy was introduced in recent decades, called Dialogue Act Markup in Several Layers (DAMSL) tag set.
According to DAMSL, each DA has a forward-looking function (such as Statement, Info-request, Question, Thanking) and a backward-looking function (such as Accept, Reject, Answer) \cite{allenCore1997}.

The DAMSL annotation includes not only the utterance-level but also segmented-utterance labelling.
However, in the emotion datasets, the utterances are not segmented.
As we can see in Figure \ref{fig:dialogue-da-emotion}, the first or fourth utterances are not segmented as two separate.
The fourth utterance could be segmented to have two dialogue act labels, for example, a statement (\textit{sd}) and a question (\textit{qy}). 
That provides very fine-grained DA classes and follows the concept of discourse compositionality.
%The emotion datasets lack of having segmented-utterances and hence annotations might incorporate some 
DAMSL distinguishes wh-question \textit{ (qw)}, yes-no question \textit{(qy)}, open-ended \textit{(qo)}, and or-question \textit{(qr)} classes, not just because these questions are syntactically distinct, but also because they have different forward functions \cite{danjur1997swbddamsl}.
For example, a \textit{yes-no question} is more likely to get a \textit{``yes"} answer than a wh-question \textit{(qw)}.
This gives an intuition that the context is provided by the answers (backward-looking function) with the questions (forward-looking function).
%It also gives an intuition that the answers follow the syntactic structure of the question, providing a context.
%In other words, discourse can be generalized for the conversation within the context of communication. 
For example, \textit{qy} is used for a question that, from a discourse perspective, expects a Yes \textit{(ny)} or No \textit{(nn)} answer.

We have investigated the annotation method and trained our neural models with the Switchboard Dialogue Act (SwDA) Corpus \cite{godfrey1992switchboard,shribergSwitch}.
% and ICSI Meeting Recorder Dialogue Act (MRDA) \cite{shriberg2004icsi}. They are annotated with the DAMSL tag set. 
The SwDA corpus is annotated with the DAMSL tag set, and it has been used for reporting and bench-marking state-of-the-art results in dialogue act recognition tasks \cite{stolcke2000dialogue,kalchbrenner2016neural,BOTHE18_525} which makes it ideal for our use case.
The Switchboard DAMSL Coders Manual\footnote{\url{https://web.stanford.edu/~jurafsky/ws97/manual.august1.html}} has more details about the dialogue act labels \cite{danjur1997swbddamsl}.

\begin{figure*}[!t]
\begin{center}
%\fbox{\parbox{6cm}{
%This is a figure with a caption.}}
\includegraphics[width=0.995\linewidth]{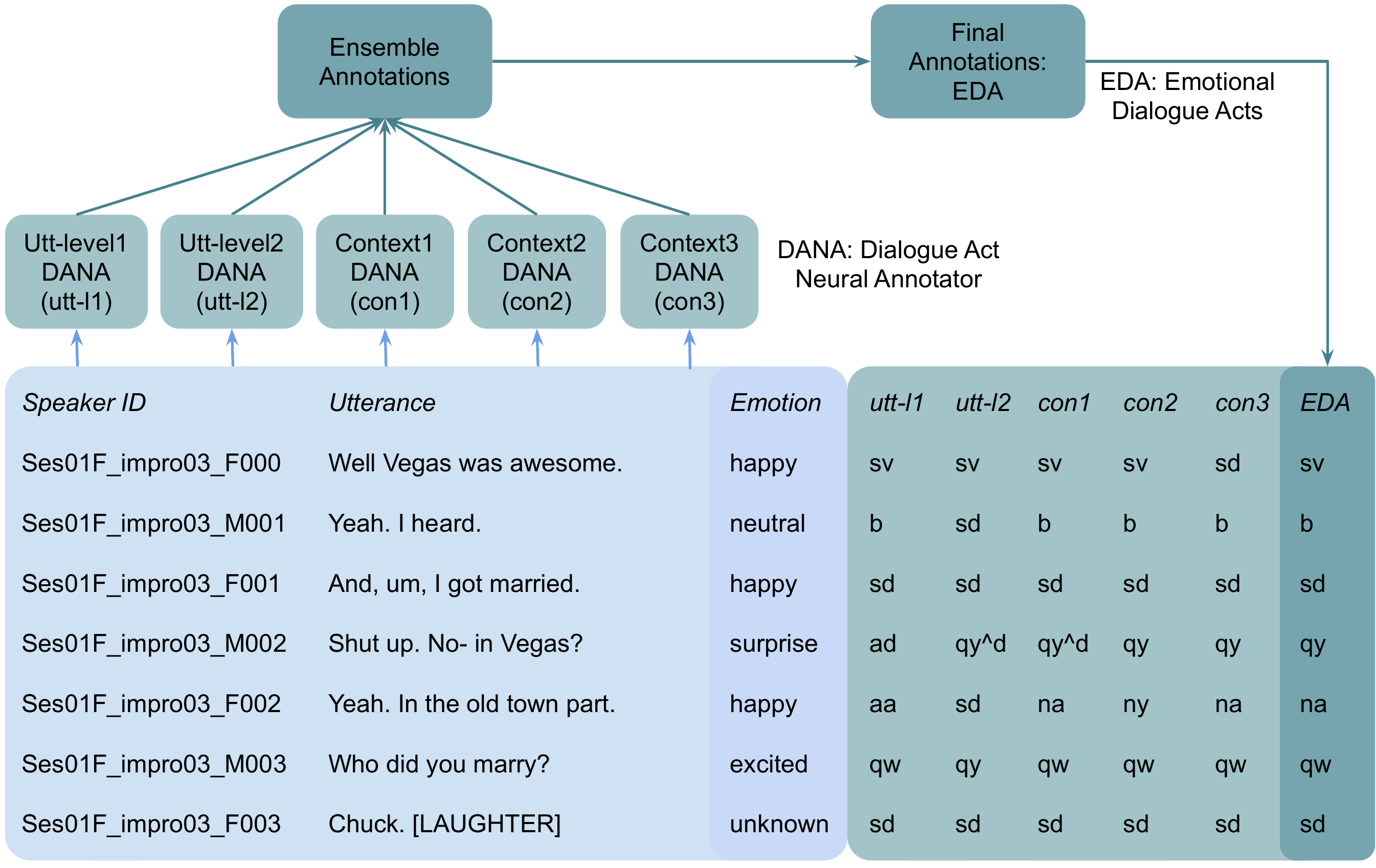} 
\caption{Setting of the annotation process of the EDAs,
% The tags of dialogue acts can be learned from Figure \ref{fig:}, 
above example utterances (with speaker identity) and emotion labels are from IEMOCAP database.}
\label{fig:pool-of-annotators}
\end{center}
\end{figure*}

\subsection{Neural Model Annotators}

We adopted the neural architectures based on Bothe et al. \shortcite{bothe2018discourse} where two variants are: a non-context model (classifying at utterance level) and a context model (recognizing the dialogue act of the current utterance given a few preceding utterances).
From a conversational analysis using dialogue acts in Bothe et al. \shortcite{bothe2018interspeech}, we learned that the preceding two utterances contribute significantly to recognizing the dialogue act of the current utterance. 
Hence, we adapt this setting for the context model and create a pool of annotators using recurrent neural networks (RNNs).
RNNs can model the contextual information in the sequence of words of an utterance, and the sequence of utterances of a dialogue.
Each word in an utterance is represented with a word embedding vector of dimension 1024.
We use the word embedding vectors from pre-trained ELMo (Embeddings from  Language  Models) embeddings\footnote{\url{https://allennlp.org/elmo}} \cite{peters2018deep} as it showed promissing performance in natural language understanding tasks \cite{wang2018glue,yang2019xlnet}.

\begin{figure}[!t]
\begin{center}
%\fbox{\parbox{6cm}{
%This is a figure with a caption.}}
\includegraphics[width=0.9\linewidth]{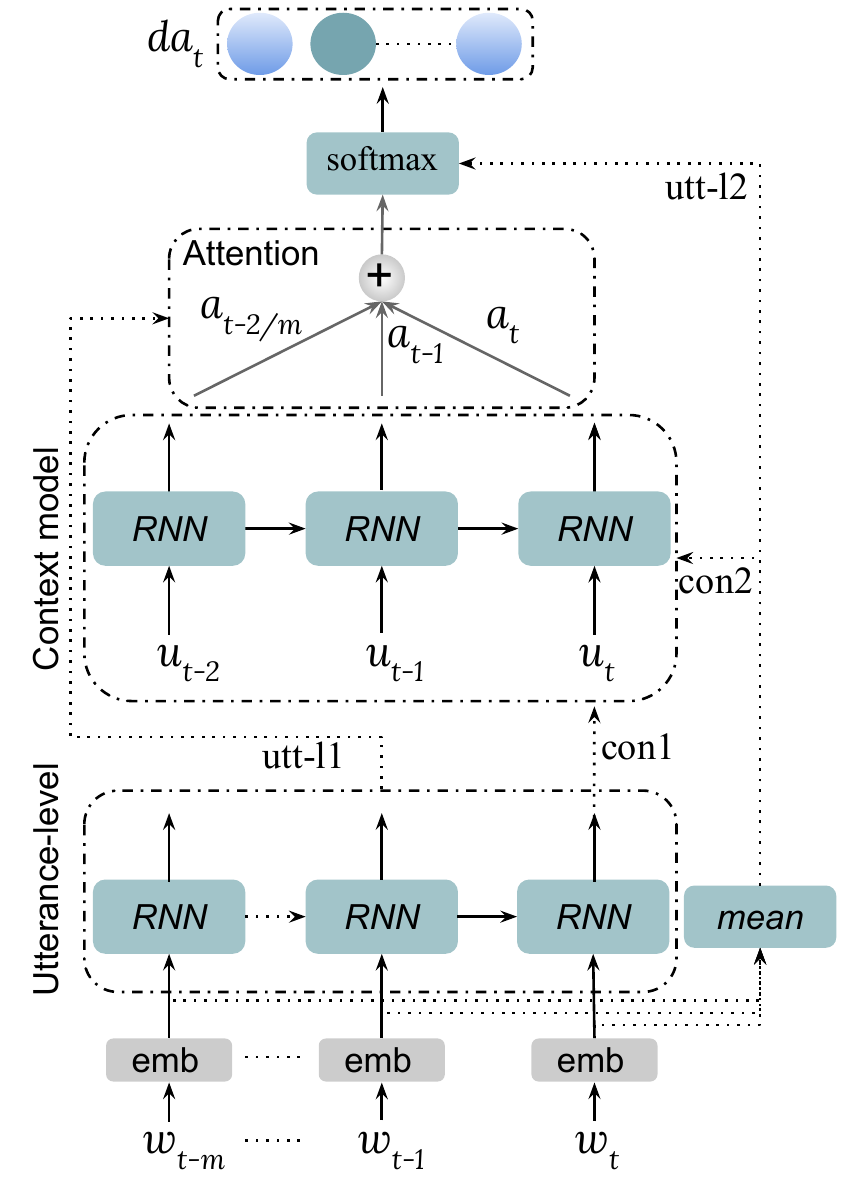} 
\caption{Recurrent neural attention architecture with the utterance-level and context-based models.}
\label{fig:neural-architecture}
\end{center}
\end{figure}

We have a pool of five neural annotators, as shown in Figure \ref{fig:pool-of-annotators}. 
Our online tool called Discourse-Wizard\footnote{\url{https://secure-robots.eu/fellows/bothe/discourse-wizard-demo/}} is available to practice automated dialogue act labelling.
In this tool, we use the same neural architectures but model-trained embeddings (while, in this work, we use pre-trained ELMo embeddings as they are better performant but computationally and size-wise expensive to be hosted in the online tool).
The annotators are:

\textbf{Utt-level-1 Dialogue Act Neural Annotator (DANA)} is an utterance-level classifier that uses word embeddings ($w$) as an input to an RNN layer, attention mechanism ($att$) and computes the probability of dialogue acts ($da$) using the \textit{softmax} function (see in Figure \ref{fig:neural-architecture}, dotted line utt-l1), formulated as:
\begin{equation}
da_t = softmax(att(RNN(w_t, w_{t-1},...,w_{t-m})))
\label{equation:RNN_first_layer}
\end{equation}
such that attention mechanism provides:
\begin{equation}
\sum_{n=0}^{n} a_{t-n} = 1
\end{equation}
This model achieved 75.13\% accuracy reported in Table \ref{table:BaselineSwDATest} on the SwDA corpus test set.

\textbf{Context-1-DANA} is a context model that uses two preceding utterances while recognizing the dialogue act of the current utterance (see context model with con1 line in Figure \ref{fig:neural-architecture}).
Context-1-DANA uses a hierarchical RNN with the first RNN layer to encode the utterance from word embeddings ($w$) as given in equation (\ref{equation:RNN_first_layer}) and the second RNN layer is provided with three utterances ($u$) (current and two preceding) composed from the first layer followed by the attention mechanism ($a$).
Finally, the \textit{softmax} function is used to compute the probability distribution, which is formulated as:
\begin{equation}
da_t = softmax(att(RNN(u_t, u_{t-1}, u_{t-2})))
\end{equation}
where $u_t$ is derived from $RNN(w_t, w_{t-1},...,w_{t-m})$.
This model achieved 77.55\% accuracy on the SwDA corpus test set see Table \ref{table:BaselineSwDATest}, it is highest performant model among the five annotators.

\textbf{Utt-level-2-DANA} is another utterance-level classifier which takes an average of the word embeddings in the input utterance and uses a feedforward neural network hidden layer (see the utt-l2 line in Figure \ref{fig:neural-architecture}, where $mean$ passed to $softmax$ directly).
Similar to the previous model, it computes the probability of dialogue acts using the \textit{softmax} function.
This model achieved 72.59\% accuracy on the test set of the SwDA corpus (see Table \ref{table:BaselineSwDATest}).

\textbf{Context-2-DANA} is another context model that uses three utterances similar to the Context-1-DANA model.
However, the utterances are composed of the mean of the word embeddings over each utterance, similar to the Utt-level-2-DANA model ($mean$ passed to context model in Figure \ref{fig:neural-architecture} with con2 line).
Hence, the Context-2-DANA model is composed of one RNN layer with three input vectors, finally topped with the \textit{softmax} function for computing the probability distribution of the dialogue acts.
This model achieved 75.97\% accuracy on the test set of the SwDA corpus (see Table \ref{table:BaselineSwDATest}).

\textbf{Context-3-DANA} is a context model that uses three utterances similar to the previous context models. 
However, the utterance representations combine both features from the Context-1 and Context-2 models (con1 and con2 together in Figure \ref{fig:neural-architecture}). 
Hence, the Context-3-DANA model combines features of almost all the previous four models to provide the recognition of the dialogue acts.
This model achieves 74.91\% accuracy on the SwDA corpus test set (in Table \ref{table:BaselineSwDATest}).

\begin{table}[!t]
\begin{center}
\begin{tabular}{llll}
Models                & Accuracy  & SC    \\ 
\hline
Utt-level-1 mdoel     & 0.751     & 0.815 \\
Context-1 mdoel       & 0.775     & 0.829 \\
Utt-level-2 mdoel     & 0.726     & 0.806 \\
Context-2 mdoel       & 0.759     & 0.823 \\
Context-3 mdoel       & 0.749     & 0.820 \\
\hline
Ensemble mdoel        & 0.778     & 0.822 \\
% \hline
\end{tabular}
\caption{Baseline validation with the SwDA test dataset.
SC: Spearman Correlation between prediction of model and ground truth.}
\label{table:BaselineSwDATest}
\end{center}
\end{table}

\begin{table}[!b]
\begin{center}
\begin{tabular}{lllll}
Stats       & AM    & CM    & BM   & NM    \\ 
\hline
% MRDA        & 43.73 & 46.66 & 3.01 & 6.60  \\
IEMOCAP     & 43.73 & 50.21 & 1.18 & 4.88  \\
MELD        & 37.07 & 51.56 & 2.20 & 9.17
\end{tabular}
\caption{Annotations Statistics of EDAs -  
AM: All Absolute Match (in \%),
CM: Context-based Models Absolute Match (in \%, matched all context models or at least two context models matched with one non-context model), 
BM: Based-on Confidence Ranking, and 
NM: No Match (in \%) (these labeled as `xx': determined in EDAs).}
\label{table:EDA_ststs}
\end{center}
\end{table}

\subsection{Ensemble of Neural Annotators}
\label{section:ensemble}

As a baseline to verify the ensemble logic, we use the SwDA test dataset where we know the ground truth labels.
Table \ref{table:BaselineSwDATest} shows the accuracy and Spearman correlation between the prediction of the model and the ground truth.
The ensemble model logic is configured in a way that it achieves an accuracy similar to or better than one of the neural annotators.
As can be seen in Table \ref{table:BaselineSwDATest}, the ensemble model achieves equivalent or a little bit better accuracy to the Context-1 model.
It is shown that the ensemble annotator performs well on the state of the art test data.
These results are also supported by the correlation scores of the respective models.
Hence, the configuration for the ensemble model that achieved the accuracy for the SwDA test dataset is explained in the following paragraph.

\begin{figure*}[!t]
\begin{center}
%\fbox{\parbox{6cm}{
%This is a figure with a caption.}}
\includegraphics[width=0.99\linewidth]{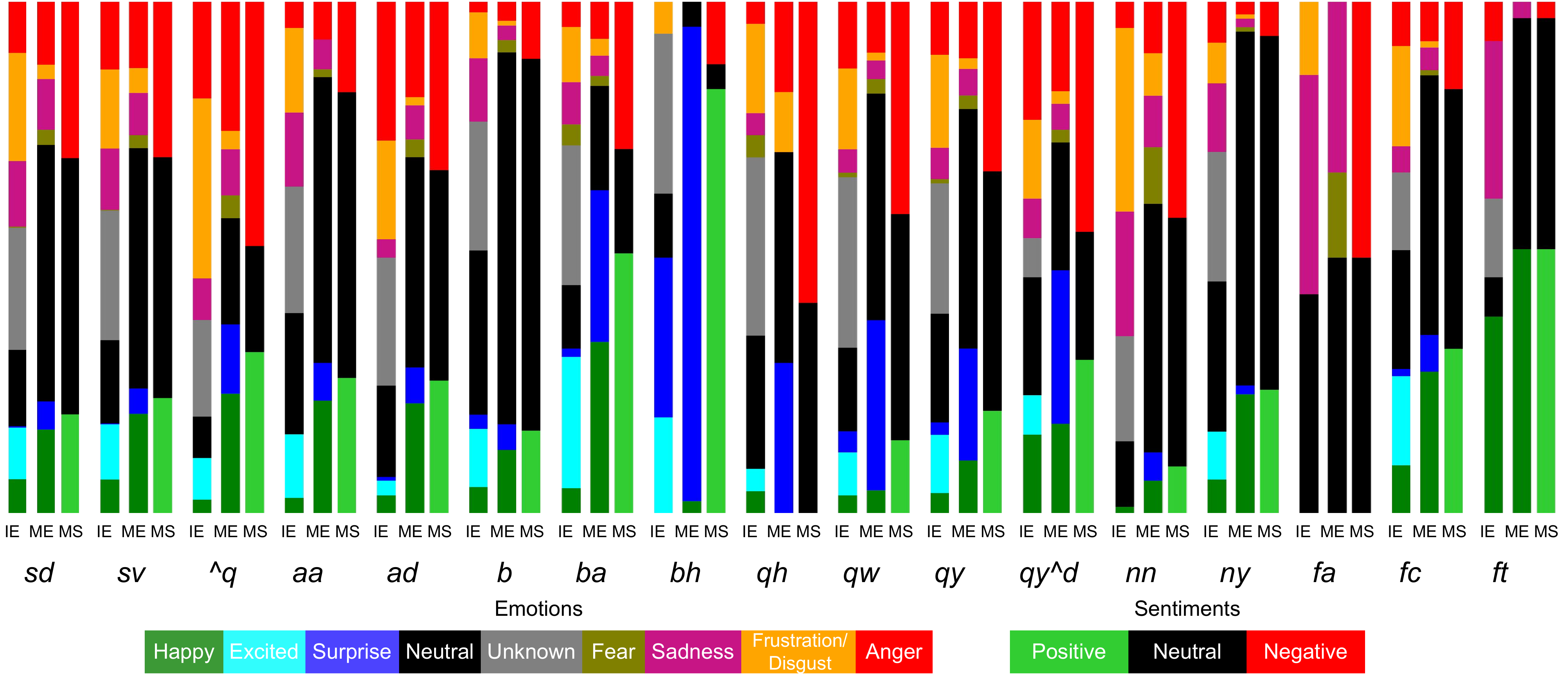} 
\caption{EDAs: Visualizing co-occurrence of utterances with respect to emotion states in the particular dialogue acts (only major and significant are shown here). IE: IEMOCAP, ME: MELD Emotion and MS: MELD Sentiment.}
\label{fig:EDA_ststistics_bars}
\end{center}
\end{figure*}

First preference is given to the labels that are perfectly matching in all the neural annotators.
In Table \ref{table:EDA_ststs}, we can see that both datasets have about 40\% of exactly matching labels over all the models (AM).
Then priority is given to the context-based models to check if the label in all context models is matching perfectly.
In case two out of three context models are correct, then it is being checked if that label is also produced by at least one of the non-context models.
Then, we allow labels to rely on these at least two context models.
As a result, about 50\% of the labels are taken based on the context models (CM).
When none of the context models is producing the same results, then we rank the labels with their respective confidence values produced as a probability distribution using the $softmax$ function.
The labels are sorted in descending order according to confidence values.
Then we check if the first three (case when one context model and both non-context models produce the same label) or at least two labels are matching, then we allow to pick that one.
There are about 1\% in IEMOCAP and 2\% in MELD (BM).

\begin{table}[!t]
\begin{center}
\begin{tabular}{llll}
DA                    & Dialogue Act names          & IEMO   & MELD   \\
\hline
sd                    & Statement-non-opinion       & 43.97  & 41.63  \\
sv                    & Statement-opinion           & 19.93  & 09.34  \\
qy                    & Yes-No-Question             & 10.3   & 12.39  \\
qw                    & Wh-Question                 &  7.26  & 6.08   \\
b                     & Acknowledge (Backchannel)   &  2.89  & 2.35   \\
ad                    & Action-directive            &  1.39  & 2.31   \\
fc                    & Conventional-closing        &  1.37  & 3.76   \\
ba                    & Appreciation or Assessment  &  1.21  & 3.72   \\
aa                    & Agree or Accept             &  0.97  & 0.50   \\
nn                    & No-Answer                   &  0.78  & 0.80   \\
ny                    & Yes-Answer                  &  0.75  & 0.88   \\
br                    & Signal-non-understanding    &  0.47  & 1.13   \\ 
\textasciicircum{}q   & Quotation                   &  0.37  & 0.81   \\
na                    & Affirmative non-yes answers &  0.25  & 0.34   \\
qh                    & Rhetorical-Question         &  0.23  & 0.12   \\
bh                    & Rhetorical Backchannel      &  0.16  & 0.30   \\
h                     & Hedge                       &  0.15  & 0.02   \\
qo                    & Open-question               &  0.14  & 0.10   \\
ft                    & Thanking                    &  0.13  & 0.23   \\
qy\textasciicircum{}d & Declarative Yes-No-Question &  0.13  & 0.29   \\
bf                    & Reformulate                 &  0.12  & 0.19   \\
fp                    & Conventional-opening        &  0.12  & 1.19   \\
fa                    & Apology                     &  0.07  & 0.04   \\
fo                    & Other Forward Function      &  0.02  & 0.05   \\
\hline
Total                 & number of utterances        & 10039  & 13708
\end{tabular}
\caption{Number of utterances per DA in the respective datasets. All values are in percentages (\%) of the total number of utterances. 
IEMO is for IEMOCAP.}
\label{table:EDA_stats_all}
\end{center}
\end{table}

\begin{table}[!t]
\begin{center}
\begin{tabular}{llll}
Metrics & $\alpha$ & $k$       & SCC  \\ 
\hline
% MRDA    & 0.553     & 0.556   & 0.636 \\
IEMOCAP & 0.553     & 0.556   & 0.636 \\
MELD    & 0.494     & 0.502   & 0.585 
\end{tabular}
\caption{Annotations Metrics of EDAs -  
$\alpha$: Krippendorff's Alpha coefficient,
$k$: Fleiss' Kappa score, and
SCC: Spearman Correlation between Context-based Models.
}
\label{table:EDA_metric_ststs}
\end{center}
\end{table}

Finally, when none the above conditions are fulfilled, we leave out the label with an unknown category.
This unknown category of the determined dialogue act is labelled with `xx' in the final annotations, and they are about 5\% in IEMOCAP and 9\% in MELD (NM).
The statistics\footnote{We are working on improving the ensemble annotation logic; hence the updated statistics will be available at the link given on the first page where corpora are available.} of the EDAs is reported in Table \ref{table:EDA_stats_all} for both corpora.
Total utterances in MELD includes training, validation and test datasets\footnote{\url{https://affective-meld.github.io/}}.

\subsection{Reliability of Neural Annotators}

\begin{table*}[!t]
\begin{center}
\begin{tabular}{llll}
EDAs                            & Utterances                    & Emotion   & Sentiment\\
\hline  
Quotation (\textasciicircum{}q) & Not after this!               & anger     & negative \\
                                & Ross, I am a human doodle!!   & anger     & negative \\
 %                               & Oh God, stop with the plan!   & anger    & negative \\
                                & No, you can't let this stop 
                                you from getting massages!      & sadness   & negative \\
                                & Oh hey! You got my 
                                parent's gift!                  & joy       & positive \\
\hline
Action-Directive (ad)           & And stop using my name!       & anger     & negative \\
%                                & Look at me!                   & anger     & negative \\
                                & Oh, let's not tell this story.& sadness   & negative \\
                                & Check it out, he's winning!   & surprise  & positive \\
                                & Yep! Grab a plate.            & joy       & positive \\
\hline
Acknowledgement/Backchannel (b) & Oh yeah, sure.                & neutral   & neutral  \\
Appreciation Backchannel (ba)   & Great.                        & joy       & positive \\
Rhetorical Backchannel (bh)     & Oh really?!                   & surprise  & positive \\
\hline
Rhetorical Question (qh)        & Oh, why is it unfair?         & surprise  & negative \\
Wh-Question (qw)                & What are you doing?           & surprise  & negative \\
                                & How are you?                  & neutral   & neutral  \\
Yes-No Question (qy)            & Did you just make that up?    & surprise  & positive \\
Declarative Yes-No Question (qy\textasciicircum{}d)  & Can't you figure that out 
                                  based on my date of birth?    & anger     & negative \\
\hline
No-Answer (nn)                  & No!                           & disgust   & negative \\
Yes-Answer (ny)                 & Yeah!                         & joy       & positive \\
\hline
Determined EDAs (xx)            &                               &           &          \\
1. (P-DA b)  b,  b, ba, fc, b   & Yeah, sure!                   & neutral   & neutral  \\
2. (P-DA sd) sv, aa, bf, sv, nn & No way!                       & surprise  & negative \\
3. (P-DA qy) aa, aa, ng, ny, nn & Um-mm, yeah right!            & surprise  & negative \\
4. (P-DA qy) aa, ar, 
\textasciicircum{}q, 
\textasciicircum{}h, nn         & Oh no-no-no, give me some 
                                  specifics.                    & anger     & negative \\
5. (P-DA fc) fc, sd, fc, sd, fp & I'm so sorry!                 & sadness   & negative \\

\end{tabular}
\caption{Examples of EDAs with annotation from the MELD dataset. 
Emotion and sentiment labels are given in the dataset, while our ensemble of models determines EDAs. P-DA: previous utterance dialogue act.}
\label{table:EDA_examples}
\end{center}
\end{table*}

The pool of neural annotators provides a fair range of annotations, 
and we checked the reliability with the following metrics \cite{mchugh2012interrater}. 
% IEMOCAP: Matches in all lists(3): 55.46\% and in context lists(2): 11.16\%, any two matches: 24.46\%, None matched: 8.92\%
% Kappa (Cohen) score between context-based predictions: 0.5598
% MELD: Matches in all lists(3): 48.25\% and in context lists(2): 11.25\%, any two matches: 25.8\%, None matched: 14.7\%
% Kappa (Cohen) score between context-based predictions: 0.4979
Krippendorff's Alpha ($\alpha$) is a reliability coefficient 
developed to measure the agreement among observers, annotators, and raters, and is often used in emotion annotation \cite{krippendorff1970estimating,wood2018emotion_annotation}.
We apply it on the five neural annotators at the nominal level of measurement 
of dialogue act categories.
$\alpha$ is computed as follows:
\begin{equation}
\alpha =1-{\frac {D_{o}}{D_{e}}}
\end{equation}
where $D_{o}$ is the observed disagreement and $D_{e}$ is the disagreement that is expected by chance. 
$\alpha=1$ means all annotators produce the same label, while $\alpha=0$ would mean none agreed on any label. 
As we can see in Table \ref{table:EDA_metric_ststs}, both datasets IEMOCAP and MELD 
produce significant inter-neural annotator agreement, 0.553 and 0.494, respectively.
However, it is a well-known problem with Kappa \cite{powers2012problem}, that dialogue acts are highly subjective and contain the unbalanced number of samples per category; still, we reach these average scores.
Hence, we decided to add one more inter-annotator metric below.

A very popular inter-annotator metric is Fleiss' Kappa score \cite{fleiss1971measuring}, also reported in Table \ref{table:EDA_metric_ststs}, which determines consistency in the ratings.
The kappa $k$ can be defined as,
\begin{equation}
    k = {\frac {\bar P -\bar P_e}{1 -\bar P_e}}
\end{equation}{}
where the denominator $1 -\bar P_e$ elicits the degree of agreement that is attainable above chance, and the numerator $\bar P -\bar P_e$ provides the degree of the agreement actually achieved above chance. 
Hence, $k = 1$ if the raters agree completely, and $k = 0$ when they do not reach any agreement.
We got 0.556 and 0.502 for IEOMOCAP and MELD, respectively, with our five neural annotators.
This indicates that the annotators are labelling the dialogue acts reliably and consistently.
We also report the Spearman's correlation between context-based models (Context-1 and Context-2), and we find a strong correlation between them (Table \ref{table:EDA_metric_ststs}).
% (as we give more weight to them)
While using the labels, we checked the absolute match between all context-based models and found that their strong correlation indicates their robustness.

\section{EDAs Analysis}
\label{section:eda-analysis}

We can see emotional dialogue act co-occurrences with respect to emotion labels in Figure \ref{fig:EDA_ststistics_bars} for both datasets.
There are sets of three bars per dialogue act in the figure, the first and second bar represents emotion labels of IEMOCAP (IE) and MELD (ME), and the third bar is for MELD sentiment (MS) labels.
MELD emotion and sentiment statistics are compelling as they are strongly correlated to each other.
The bars contain the normalized number of utterances for emotion labels concerning the total number of utterances for that particular dialogue act category.
The statements without-opinion (\textit{sd}) and with-opinion (\textit{sv}) contain utterances with almost all emotions. 
Many neutral utterances are spanning over all the dialogue acts.

Quotation (\textit{\textasciicircum{}q}) dialogue acts, on the other hand, are mostly used with `Anger' and `Frustration' (in case of IEMOCAP), but some utterances with `Joy' or `Sadness' as well (see examples in Table \ref{table:EDA_examples}). 
Action Directive (\textit{ad}) dialogue act utterances, which are usually orders, frequently occur with `Anger' or `Frustration' although many also with the `Happy' emotion in case of the MELD dataset.
Acknowledgements (\textit{b}) are mostly used with positive or neutral sentiment, however, Appreciation (\textit{ba}) and Rhetorical (\textit{bh}) backchannels often occur with a greater number in `Surprise', `Joy' and/or with `Excited' (in case of IEMOCAP).
Questions (\textit{qh, qw, qy and qy\textasciicircum{}d}) are mostly asked with emotions `Surprise', `Excited', `Frustration' or `Disgust' (in case of MELD), and many are neutral.
No-answers (\textit{nn}) are mostly `Sad' or `Frustrated' as compared to yes-answers (\textit{ny}).
Forward-functions such as Apology (\textit{fa}) are mostly used with `Sadness'
whereas Thanking (\textit{ft}) and Conventional-closing or -opening (\textit{fc} or \textit{fp}) are usually with `Joy' or `Excited'.

We also noticed that both datasets exhibit a similar relation between dialogue act and emotion.
The dialogue act annotation is based on the given transcripts; however, the emotional expressions are better perceived with audio or video \cite{busso2008iemocap,lakomkin2019incorporating}. 
We report some examples where we mark the utterances with a determined label (`xx') in the last row of Table \ref{table:EDA_examples}.
They are left out from the final annotation (labeled as determined EDA `xx') because of not fulfilling the conditions explained in Section \ref{section:ensemble}
It is also interesting to see the previous utterance dialogue acts (P-DA) of those skipped utterances, and the sequence of the labels can be followed from Figure \ref{fig:pool-of-annotators} (utt-l1, utt-l2, con1, con2, con3).

In the first example, the previous utterance was \textit{b}, and three DANA models produced labels of the current utterance as \textit{b}, but it is skipped because the confidence values were not sufficient to bring it as a final label.
The second utterance can be challenging even for humans to decide with any of the dialogue acts.  
However, the third and fourth utterances are followed by a yes-no question (\textit{qy}), and hence, we can see in the third example, that context models tried their best to at least perceive it as an answer (\textit{ng, ny, nn}).

The last utterance, ``I'm so sorry!", has different results by all the five annotators.
Similar apology phrases are mostly found with `Sadness' emotion label, and the correct dialogue act is Apology (\textit{fa}).
However, they are placed either in the \textit{sd} or in \textit{ba} dialogue act category.
This mostly occurs due to less number of examples in the dialogue act categories like \textit{fa} or \textit{ft}.
See Table \ref{table:confused_edas}, where the EDAs are either wrongly determined or confused by all the annotators. 
It is essential that the context-based models are looking into the previous utterances; hence, the utterance ``Thank you." can be treated as backchannel acknowledgement (\textit{ba}).
Hence, we believe that with human annotator's help, those labels of the utterances can be corrected with minimal efforts.

\begin{table}[!t]
\begin{center}
\begin{tabular}{llll}
Utterances                & Emotion  & Annotators & EDA  \\ 
\hline
I'm sorry. & sadness & ba,sd,fc,fc,fc & fc  \\
Dude, I am sorry  &  & &  \\
about what I said! & sadness & sd,fa,\textasciicircum{}q,sd,sd & sd \\
Sorry, Pheebs. & sadness & fc,fa,ad,fa,ad & ad \\
I am so sorry... & sadness & sd,sd,\textasciicircum{}q,sd,sd & sd \\
Thank you.  & neutral & fc,fc,ba,ft,ba & ba \\
Thank you we're  &  &  &  \\
so excited. & joy & fc,sd,fc,fc,fc & fc \\
Nice,  thank you. & joy & fc,fc,ba,ft,ba & ba \\
\hline
\end{tabular}
\caption{Examples of wrongly determined (or confused) EDAs with annotation from the MELD dataset.}
\label{table:confused_edas}
\end{center}
\end{table}

\section{Conclusion and Future Work}

In this work, we presented a method to extend conversational multi-modal emotion datasets with dialogue act labels.
The ensemble model of the neural annotators was tested on the Switchboard Dialogue Acts corpus test set to validate its performance. 
We successfully annotated two well-established emotion datasets: IEMOCAP and MELD, which we labelled with dialogue acts and made them publicly available for further study and research.
As a first insight, we found that many of the dialogue acts and emotion labels follow certain relations. 
These relations can be useful to learn about the emotional behaviours with dialogue acts to build a natural dialogue system and for more in-depth conversational analysis.
The association between dialogue act and emotion labels is highly subjective.
However, the conversational agent might benefit in generating an appropriate response when considering both emotional states and dialogue acts in the utterances.

In future work, we foresee the human in the loop for the annotation process along with a pool of automated neural annotators.
Robust annotations can be achieved with minimal human effort and supervision, for example, observing and correcting the final labels produced by ensemble output labels from the neural annotators.
The human-annotator might also help to achieve segmented-utterance labelling of the dialogue acts.
We also plan to use these corpora for conversational analysis to infer interactive behaviours of the emotional states with respect to dialogue acts.
In our recent work, where we used dialogue acts to build a dialogue system for a social robot, we find this study and datasets very helpful.
For example, we can extend our robotic conversational system to consider emotion as an added linguistic feature to produce a more natural interaction.

\section{Acknowledgements}
We would like to acknowledge funding from the  European Union's Horizon 2020 research and innovation programme under the Marie Sklodowska Curie grant agreement No 642667 (SECURE).

\balance

% \nocite{*}
\section{Bibliographical References}
\label{main:ref}

\bibliographystyle{lrec}
\bibliography{thesis}

\end{document}